\documentclass{article}
\usepackage{ijcai23}

\usepackage{times}
\usepackage{soul}
\usepackage{url}
\usepackage[hidelinks]{hyperref}
\usepackage[utf8]{inputenc}
\usepackage[small]{caption}
\usepackage{graphicx}
\usepackage{amssymb}
\usepackage{bm}
\usepackage{subfigure}
\usepackage{amsmath}
\usepackage{amsthm}
\usepackage{booktabs}
\usepackage{algorithm}
\usepackage{algorithmic}
\usepackage[switch]{lineno}

\title{TaiChi Action Capture and Performance Analysis\\
 with Multi-view RGB Cameras}

\author{
Jianwei Li\and
Siyu Mo\And
Yanfei Shen
\affiliations
Beijing Sport University\\
\emails
\{jianwei,msy,syf\}@bsu.edu.cn,
}

\begin{document}

\maketitle

\begin{abstract}
Recent advances in computer vision and deep learning have influenced the field of sports performance analysis for researchers to track and reconstruct freely moving humans without any marker attachment. However, there are few works for vision-based motion capture and intelligent analysis for professional TaiChi movement. In this paper, we propose a framework for TaiChi performance capture and analysis with multi-view geometry and artificial intelligence technology. The main innovative work is as follows: 1) A multi-camera system suitable for TaiChi motion capture is built and the multi-view TaiChi data is collected and processed; 2) A combination of traditional visual method and implicit neural radiance field is proposed to achieve sparse 3D skeleton fusion and dense 3D surface reconstruction. 3) The normalization modeling of movement sequences is carried out based on motion transfer, so as to realize TaiChi performance analysis for different groups. We have carried out evaluation experiments, and the experimental results have shown the efficiency of our method.
\end{abstract}

\section{Introduction}
\label{sec:0}

Human motion capture (Mocap) is usually used to obtain 3D human movement information in proactive health care and intelligent sports. Sports performance analysis and evaluation can improve athletes' competitive ability or promote public scientific fitness. The widely used inertial and optical motion capture systems can track and record human movement well, but need to bind sensors or paste marks on human body, which may affect human movement. Moreover, most current optical and inertial motion capture systems are expensive, and how to stick the markers also requires certain professional knowledge. Visual motion capture methods use cameras to non-invasively capture human motion images and then obtain the motion data through human pose estimation (HPE) and 3D reconstruction. Vision-based human action analysis is an important research topic in computer vision, and in recent years it has been widely applied in intelligent sports. Accurate 3D human motion modeling is the prerequisite for reliable human motion analysis.

Human action recognition (HAR) and action quality assessment (AQA) are two tasks of performance analysis in intelligent sports. The former aims to identify the action classification, while the latter aims to automatically quantify the performance of the action or to score its performance. Traditional methods for action analysis are mainly based on artificial design features, and compare the action sequences by estimating the distance error or dynamic time warping. Deep-learning methods use the deep network to directly learn action features and have shown more powerful performance.
Generally speaking, deep-learning methods consist of video-based methods and skeleton-based methods.
Algorithms based on video generally extract features directly from images, such as C3D \cite{2020What}, I3D \cite{2017Quo}, and TSN \cite{2018S3D} and Pseudo3D \cite{qiu2017learning}, and then extract time domain features by LSTM, pooling, and so on. The finally score prediction is performed by a fully connected neural network.
Skeleton-based methods first detect the human skeleton in the images or video, and then model the correlation information between human joints, so as to realize human motion modeling and motion quality evaluation.

At present vision-based performance analysis is mature in motion recognition and has made remarkable progress, but the performance of action quality assessment in sports motion scoring and intelligent sports training is still lower than the current application needs.
Many studies on human movement evaluation with computer vision have been proposed, however most of them only select a few relatively simple fitness or clinical rehabilitation movements to identify and evaluate.
As the current mainstream method, deep learning needs large-scale human motion dataset to train a better model, which limits their effects on sports action analysis.
Although some human professional sports datasets have been presented in recent years, however most of them are RGB images or videos collected from the Internet, such as AQA-7\cite{parmar2019action} and Yoga-82 \cite{verma2020yoga}. The performance of sports scoring or quality evaluation is still below the current application requirements.

\begin{figure}
\begin{center}
 \includegraphics[height=3.6cm]{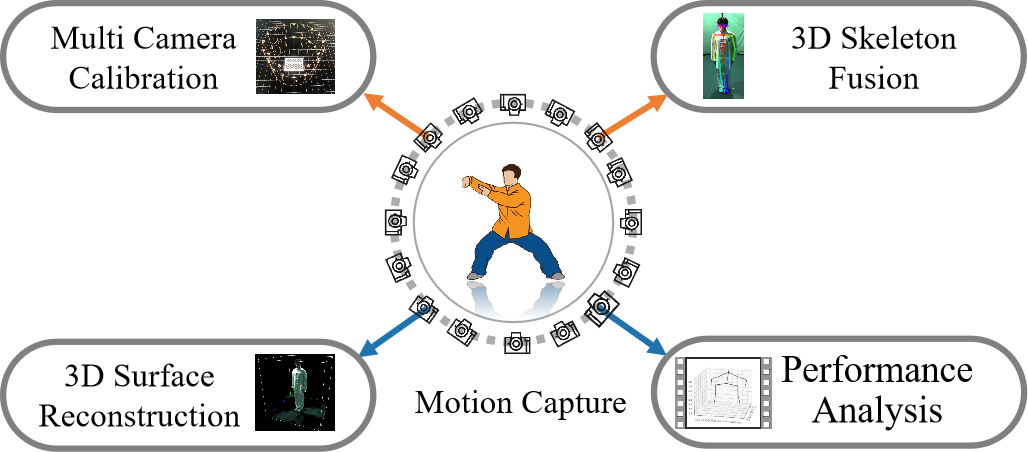}
\end{center}
  \caption{The system of TaiChi performance capture and analysis.}
\label{fig:pipe}
\end{figure}

According to above analyses, existing studies focus more on the recognitions of regular actions or assessments of competitive sports, and lack of 3D action dataset. As a complement to above work, we focus on how to capture and intelligently assess the quality of TaiChi actions.
In summary the main {\bf contributions} of this paper are the followings:
\begin{enumerate}
\item A professional TaiChi dataset consists of 23,232 action samples captured through multi-view cameras;
\item An effective 3D human modelling framework with multi-camera calibration, 3D skeleton fusion and 3D surface reconstruction;
\item A normalized modeling method for skeleton sequences based on motion transfer to analyse TaiChi performance from different groups.
\end{enumerate}
\section{Related Work}
\textbf{Vision-based motion capture.} Human Mocap system based on vision technology can obtain 3D movement information non-invasively, and is gradually applied in the field of sports performance analysis.
The type of human motion description includes human skeleton model (e.g., MPII \cite{andriluka14cvpr}), human parametric model (e.g., SMPL \cite{loper2015smpl}) and dense shape model (e.g., HumanNeRF \cite{weng2022humannerf}). Among them, the skeleton model of human body describes the non-rigid motion of 3D surface with high degrees of freedom as surface motion driven by the dynamic chain. As a structured representation of human pose, skeleton model can conveniently and effectively represent quantitative information of human movements, and is widely used in action analysis.
According to the number of viewpoints, visual Mocap systems can be divided into single-view system and multi-view system. The single-view system generally uses a single camera to capture human motion from a fixed perspective, while the multi-view system obtains human motion images from multiple perspectives based on the multi-camera system. The main challenge of single-view method is the problems of occlusion and depth uncertainty. Compared with the single-view method, the multi-view method can provide multi-view information, which can alleviate the occlusion problems and better restore 3D human posture.
Imocap\cite{dong2020motion} can capture human motion from multiple Internet videos, and open a new direction for 3D HPE. DeepMultiCap\cite{zheng2021deepmulticap} uses the pixel alignment implicit function based on the parameterized model to reconstruct the invisible region response to the severe occlusion problem in the close-range interaction scene, and captures geometric details of human surface over time based on the attention module. Multi-view video data can contain more temporal and spatial information, but human posture may vary dramatically in continuous frames of video data, so how to integrate the data effectively remains to be solved.
Currently, mainstream deep learning methods often require a large number of labeled data, which increases the difficulty of model training for sports action.

\textbf{Vision-based human motion datasets.}
NTU RGB+D \cite{NTURGBD120} is so far the largest Kinect-based action dataset collected from 106 distinct subjects and contains more than 114 thousand video samples and 8 million frames. The dataset contains 120 different action classes including daily, mutual, and health-related activities.
Human 3.6M \cite{Ionescu2014Human3} is another large dataset with 3.6 million human poses and corresponding images. There are 11 subjects and 17 action scenes, and the data is made up of four digital cameras, one time sensor and ten motion cameras. 
UCF-sport \cite{2014Action} is the first sports action dataset, and contains close to 200 action video sequences collected from various sports which are typically featured on broadcast television channels such as BBC and ESPN.
Since then a number of sports motion datasets \cite{li2018resound,shao2020finegym,verma2020yoga} used for action recognition have emerged.
FineGym \cite{shao2020finegym} provides coarse-to-fine annotations both temporally and semantically for gymnastics videos. There are three levels of categorical labels, and the temporal dimension is also divided into two levels, i.e., actions and sub-actions.
UMONS-TAICHI \cite{tits2018umons} includes 2,200 sequences of 13 classes (relative to different Taijiquan techniques) performed by 12 participants of different levels of expertise.
Fitness-AQA \cite{parmar2022domain} is a new exercise dataset comprising of three exercises (\emph{BackSquat}, \emph{BarbellRow} and \emph{OverheadPress}), has been annotated by expert trainers for multiple crucial and typically occurring exercise errors.
At present, most sports action datasets used for performance analysis are competition data based on publicly available RGB images or videos, such as public FSD-10\cite{liu2020fsd} and Finediving \cite{xu2022finediving}, but few of them have multi-view 3D skeleton poses.

\textbf{Vision-based action analysis.}
With the development of intelligent sports and computer vision, many action analysis methods for sports have been gradually proposed in recent years. Deep learning is currently the mainstream method for vision-based action analysis, where the most widely used models are RNNs, CNNs, GCNs and Transform-based. According to the type of input data, there are mainly image-based \cite{duan2022revisiting,parmar2019and} and skeleton-based \cite{yan2018spatial,pan2019action} methods. ScoringNet \cite{li2018scoringnet} and SwingNet \cite{mcnally2019golfdb} are based on images, and support fine-grained action classification and action scoring. These methods focus on the visual activity information of the whole scene including the performer's body and background, but may tend to ignore the motion relationship within the human skeleton joints.
Skeleton-based methods generally begin by extracting human skeleton, and then conduct spatio-temporal modeling on the association information between skeleton joints. For example, the joint relational graph method proposed by Pan et al. \cite{pan2019action} models the conventional motion and motion difference between different parts of human body according to the joint common module and joint difference module respectively.
HDVR \cite{hu2021unsupervised} proposes a hierarchical dance video recognition framework by estimating 3D human pose from the corresponding 2D human pose sequences. For competitive gymnastics, SportsCap \cite{chen2021sportscap} uses ST-GCN \cite{yan2018spatial} method to predict a fine-grained semantic action attributes, and adopts a semantic attribute mapping block to assemble various correlated action attributes into a high-level action label for the overall detailed understanding of the whole movement.
In recent years, vision-based HAR methods are mature and have made remarkable progress, and AQA technologies have also been developed gradually. However, human body is often in self-occlusion with large folding or bending in sports, the performance of existing methods in rehabilitation training and sports scoring is still lower than current application requirements. The recognition accuracy of uncommon or highly similar human motion is still limited, and how to effectively model and analyze human motion with challenging situation, such as complex movements and view change, needs to be further studied.

\section{Methods}
For TaiChi performance capture and analysis, we design a non-invasive system with multi-view geometry and artificial intelligence technology.
\begin{figure}
\begin{center}
 \includegraphics[height=3.2cm]{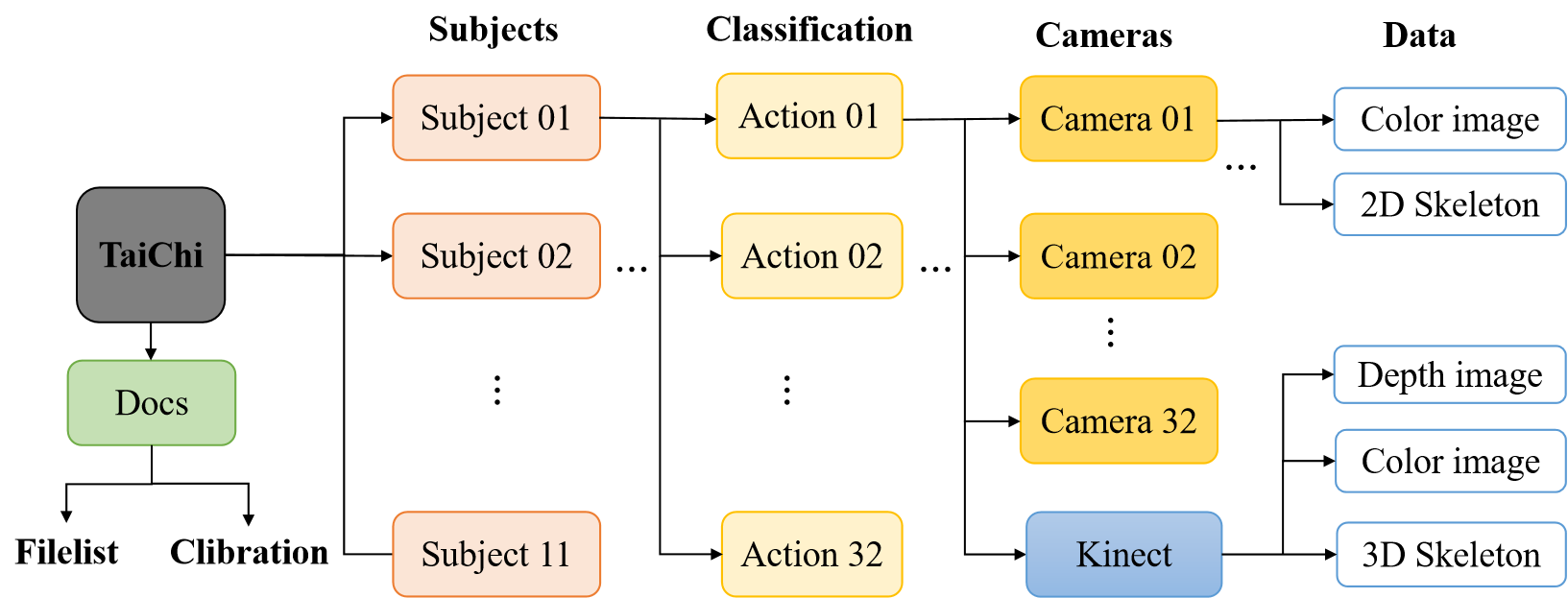}
\end{center}
  \caption{The organization of TaiChi data.}
\label{fig:org}
\end{figure}

\subsection{Experimental Setup}
Since there has a lot of body rotation in TaiChi movements, we set up a ring array multi-camera system to realize a better motion capture.
The installation bracket of the system is a positive 16-sided shape with a diameter of 450 \emph{cm} and a height of 250 \emph{cm}, with a total of 16 columns.
Each two RGB cameras (2448$\times$2048p) from the FLIR company are installed on each column. There are 32 cameras in total, 16 of them are 100 \emph{cm} from the ground and the rest of them are 200 \emph{cm} from the ground. Cameras on the top are tilted down about 20 degrees, and cameras on the bottom are tilted down about 10 degrees. Each four cameras are connected to a server, and the 8 servers are networked through a 10 GB router. All cameras are synchronously controlled through a special trigger device.
Data process and experiments are conducted on the PyTorch deep-learning framework on a standard desktop PC with 11 GB 1080Ti GPUs.

\subsection{System Composition}

As shown in Figure \ref{fig:pipe}, the proposed system mainly contains five modules:
\begin{itemize}

\item Multi-camera calibration. Before motion capture, the system is calibrated to obtain the internal reference matrix of each camera and the pose relationship between multi cameras.

\item Motion capture. The 32 high frame rate RGB cameras through 8 servers are synchronously controlled to capture and store the high-definition TaiChi motion data.

\item 3D skeleton fusion. The 2D skeletons are obtained by the HPE method from each RGB image and then fused into 3D human skeleton by multi-visual geometry matching.

\item 3D surface reconstruction. The 3D human surface model is reconstructed with multi-view images through camera poses estimation and neural radiance field rendering.

\item Performance analysis. The skeleton sequences of different subjects are transferred to a standard model to eliminate individual appearance differences, and then TaiChi action quality is compared and evaluated by comparing the trajectory and angle changes of the re-targeted skeletons.
\end{itemize}

\section{TaiChi Performance Capture}
\label{sec:2}

\subsection{Multi-view Data Organization}
\begin{figure}
\begin{center}
 \includegraphics[height=3.2cm]{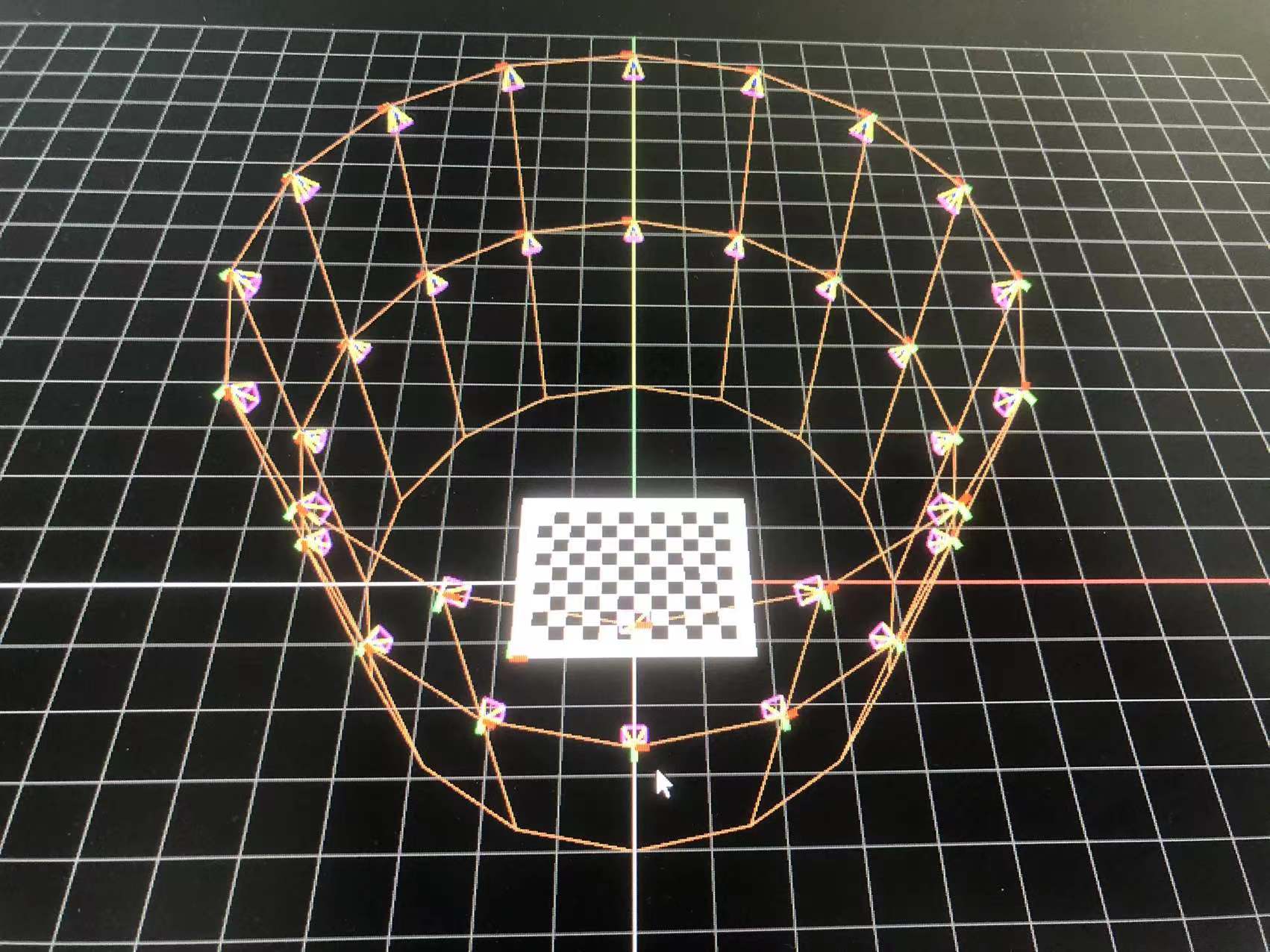}
\end{center}
  \caption{Visual simulation of our multi-camera calibration.}
\label{fig:calib}
\end{figure}

Figure \ref{fig:org} shows the organization of multi-view TaiChi action data, which contains 23,232 action samples, including each sample’s RGB image, depth image, 2D skeleton and 3D skeleton data. Each TaiChi action sample is captured by 32 RGB cameras from 32 different views and a RGB-D camera (Kinect Azure) from the front view simultaneously. During TaiChi data acquisition, 11 subjects (3 female and 8 male) have performed the 24-form TaiChi actions same as TaiChi-24 \cite{li2022tai}. Each action sample is manually segmented and labeled with category and action quality. The 2D skeletons are computed through Openpose \cite{cao2017realtime} algorithm, while the 3D skeletons are obtained from Kinect Azure SDK. 

\subsection{Multi-camera Calibration}

To get the pose relationship of the 32 RGB cameras, we design a multi-camera calibration tool by the 2D planar checkerboard calibration method \cite{zhang2000flexible}. Figure \ref{fig:calib} shows a visual simulation of our multi-camera calibration process. The 2D checkerboard is located in the center of the multi-camera system, and its orientation and pitch angle are changed uniformly. More than 100 checkerboard images are selected in each calibration. The grids in the checkerboard are 10 $\times$ 15, and the actual side length of each grid is 5 \emph{cm}. In the process of data acquisition and processing, we have carried out 10 times of calibration. Based on the camera projection model, we construct a minimization objective function with re-projection error to solve the camera parameters: 

\begin{equation}\label{}
min\sum\|\mathbf{P}\mathbf{X}_i-\mathbf{x}_i\|^2,
\end{equation}
where $\mathbf{x}$ is the feature in RGB image, $\mathbf{X}$ is the of in the checkerboard, and $\mathbf{P}=\mathbf{K}[\mathbf{R}|\mathbf{t}]$ is the camera projection matrix. $\mathrm{\mathbf{K}}$ is the $i$-th camera internal reference matrix, and $[\mathrm{\mathbf{R}}_i|\mathrm{\mathbf{t}}_i]$ is the $i$-th camera external reference matrix. In order to further improve the accuracy of calibration, bundle-adjustment (BA) \cite{triggs2000bundle} is used for optimization.

\subsection{3D Skeleton Fusion}
2D skeletons estimated from a single view RGB image often have the problem of occlusion, which will affect the accuracy of performance analysis. Therefore, we make 3D skeleton fusion by direct linear transformation (DLT) \cite{adbel1971direct} algorithm. The main calculation process is shown in the following formula:
\begin{figure}
\begin{center}
 \includegraphics[height=3.6cm]{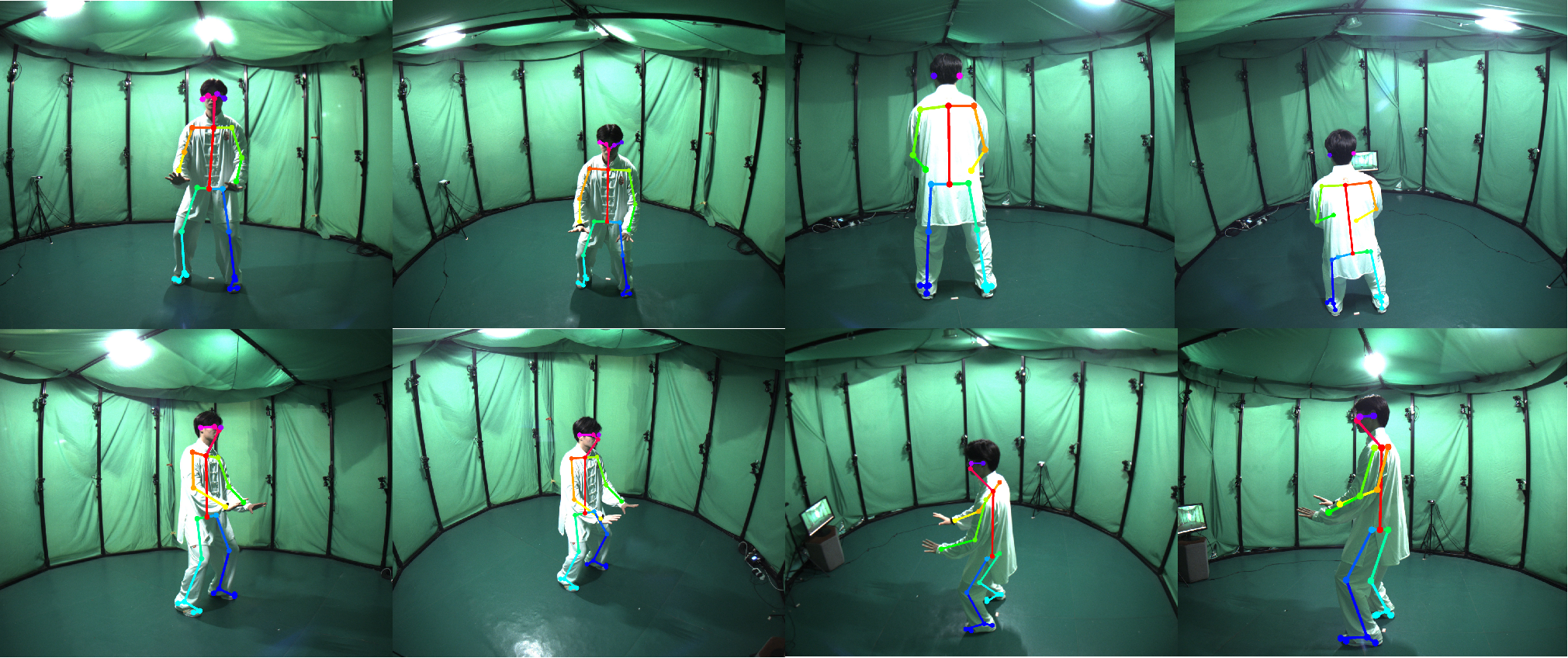}
\end{center}
  \caption{Example of 3D skeleton fusion results from 8 different views.}
\label{fig:skeleton}
\end{figure}

\begin{equation}\label{3d}
\mathrm{\mathbf{s}}_{i}=\mathrm{\mathbf{K}}_i[\mathrm{\mathbf{R}}_i|\mathrm{\mathbf{t}}_i]\mathrm{\mathbf{S}},i\in(1,m)
\end{equation}
where $\mathrm{\mathbf{s}}_i=\{s_1,s_2,...s_N\}$ is the 2D skeleton with $N$ joints in the $i$-th camera view, $\mathrm{\mathbf{S}}=\{S_1,S_2,...S_N\}$ is the corresponding 3D skeleton, and $m$ is the number of fused camera views. Figure \ref{fig:skeleton} shows an example of 3D skeleton fusion results (rendered on 2D images) for a subject from 8 different views.

\subsection{3D Surface Reconstruction}

Considering the excellent modeling and rendering capabilities of neural radiation fields (NeRFs), we realize 3D human surface reconstruction by joint using the traditional Colmap \cite{schonberger2016structure} and deep-learning based Instant NeRF \cite{muller2022instant}: 1) Firstly, we detect and extract SIFT features from each input image; 2) And then the positions of the multi cameras are estimated by feature matching; 3) Finally, data conversion is performed for NeRF rendering. Given a 3D point and a viewing direction $d\in \mathbb{R}^3$, NeRF estimates RGB color values and density ($c,\sigma$) that are then accumulated via quadrature to calculate the expected color of each camera ray:

\begin{equation}\label{}
C(r)=\int^{t_f}_{t_n}exp(-\int^{t}_{t_n}\sigma(s)ds)\sigma(t)c(t,d)dt,
\end{equation}
where $t_f$ and $t_n$ define the near and far bounds, and the camera ray is indicated as $r(t)=o+td$.

\begin{figure}
\begin{center}
 \includegraphics[height=6.9cm]{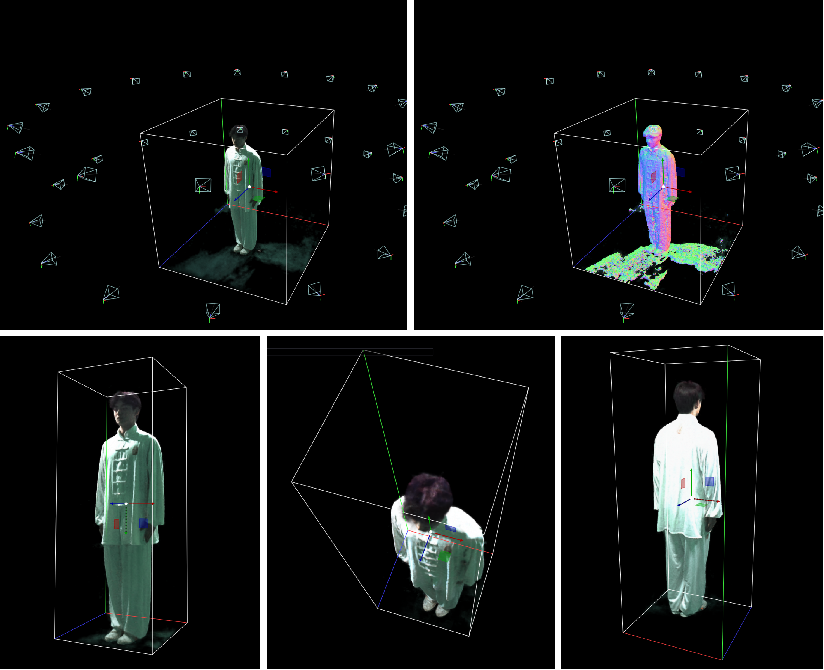}
\end{center}
  \caption{3D human surface reconstruction with NeRFs.}
\label{fig:nerf}
\end{figure}

In order to speed up the signed distance function (SDF) training, 3D training positions are uniformly sampled and computed to the triangle mesh.
Figure \ref{fig:nerf} shows the 3D surface reconstruction results of a subject with Instant NeRF. The top are the overall relationship between the camera poses and human models, while the bottom are the reconstruction details from the front, top and back views respectively.
\section{TaiChi Performance Analysis}

\subsection{Data Precision Analysis}

\begin{figure*}
\begin{center}
\subfigure[The wearing positions of the PNS sensor]{
\begin{minipage}[t]{0.45\linewidth}
\centering
\includegraphics[height=3.2cm]{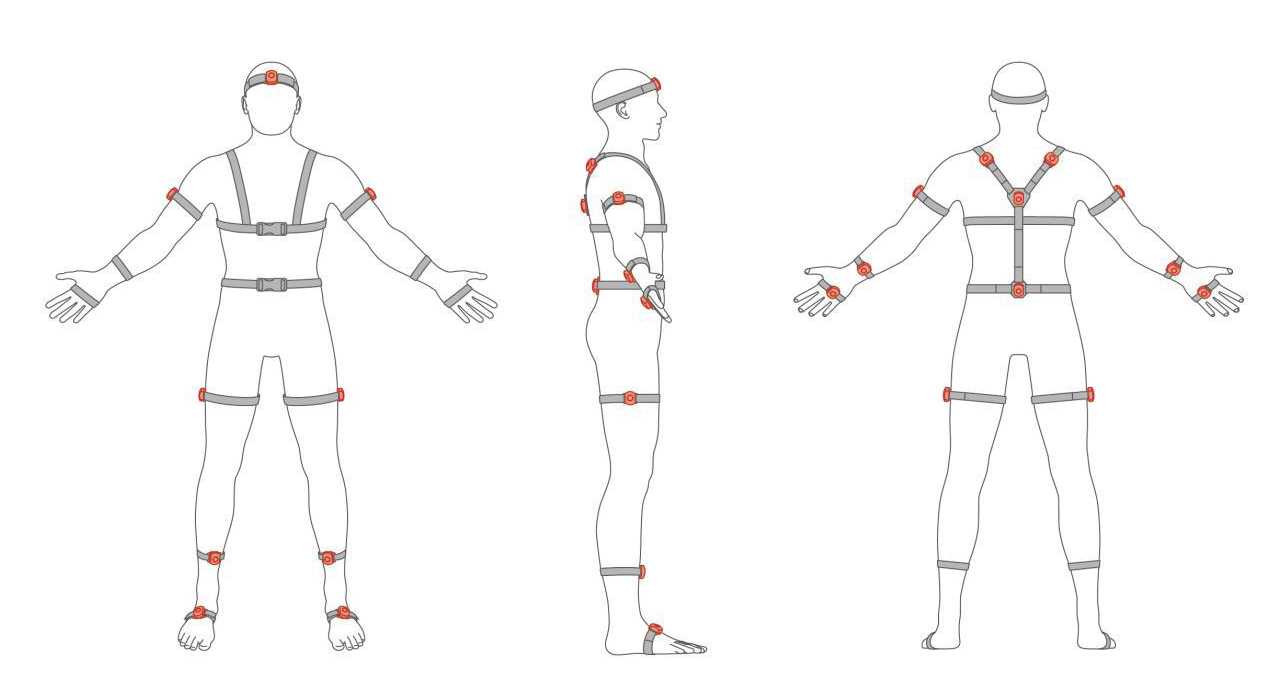}
\end{minipage}
}
\subfigure[The joints extracted with Openpose algorithm]{
\begin{minipage}[t]{0.45\linewidth}
\centering
\includegraphics[height=3.1cm]{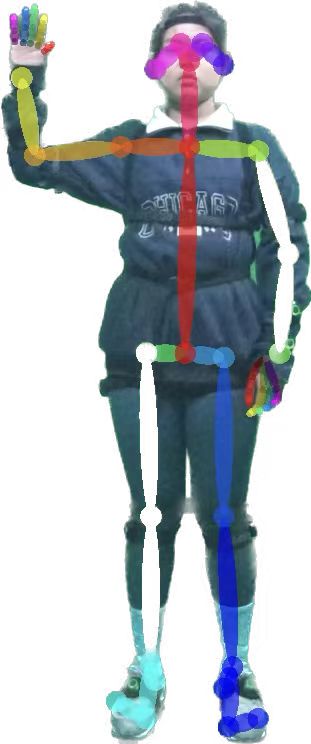}
\end{minipage}
}
\end{center}
  \caption{Comparison of the wearing positions of IMU and human skeleton joints extracted from our visual Mocap system.}
\label{fig:imu}
\end{figure*}
Accurate recovery of 3D human movement information is the premise of reliable analysis of human movement.
In order to verify the accuracy of our multi-camera system with IMU-based motion capture system, we conduct experiments using these two systems in time and space synchronization. A subject has wore the IMU equipments (Perception Neuron Studio, PNS) and performed three upper limb exercises (\emph{lateral hand lift}, \emph{left punch} and \emph{right punch}) and three lower limb exercises (\emph{left knee lift}, \emph{right knee lift} and \emph{lunging squat}) when the frame rates of the multi-camera system are 30 fps and 60 fps respectively.
Figure \ref{fig:imu} shows the wearing position of the PNS sensors and human skeleton nodes (Body-25) extracted by Openpose. It can be seen that there has a position deviation for the joints obtained by these two Mocap systems.

\begin{figure}
\begin{center}
 \includegraphics[height=4.8cm]{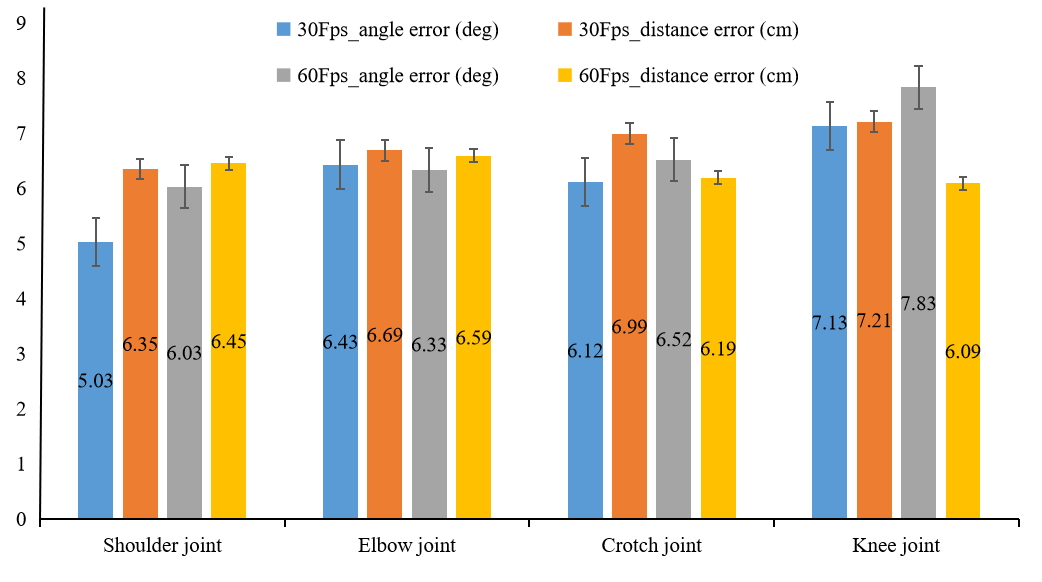}
\end{center}
  \caption{Comparison of multi-camera and IMU-based Mocap system by angle error (\emph{deg}) and distance error (\emph{cm}) of key joints.}
\label{fig:evaluation}
\end{figure}

We have compared and analyzed the Mocap data by the coordinate information of \emph{shoulder}, \emph{elbow}, \emph{hip} and \emph{knee} joints.
Figure \ref{fig:evaluation} shows the MSE error between our multi-camera system and IMU-based Mocap system. The average angle error and distance error are 6.18 \emph{deg} and 6.81 \emph{cm} respectively in 30 fps. The average angle error and distance error are 6.68 \emph{deg} and 6.41 \emph{cm} respectively in 60 fps. The main reason for the error lies in the deviation between the paste positions of the PNS sensor and skeleton nodes obtained by visual estimation (as shown in Figure \ref{fig:imu}).

\subsection{Performance Analysis}

For TaiChi performance analysis, we combine 32 camera views from 16 orientations into 16 stereo pairs to obtain 3D skeletons.
3D skeleton of each camera pair also is computed with 3D reconstruction module of Openpose. The quantization accuracies of TaiChi action recognition and assessment have been discussed in \cite{li2022tai}. Therefore, we mainly conduct performance analysis by comparing the movements between the coach and the students. To avoid the influence of individual differences, the action skeleton sequences from the coach and students are uniformly transferred to a standard virtual human model for comparison.

\begin{figure}
\begin{center}
 \includegraphics[height=2.6cm]{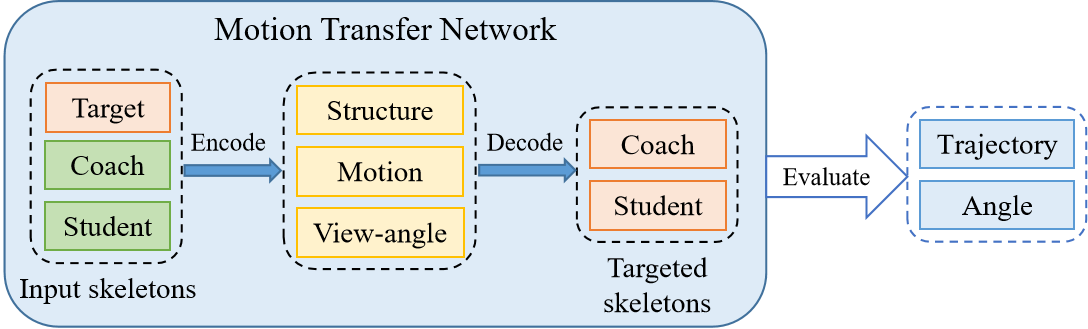}
\end{center}
  \caption{The flowchart of performance analysis with motion transfer network.}
\label{fig:net}
\end{figure}

The flowchart of performance analysis with the motion transfer network is shown in Figure \ref{fig:net}. The model decomposes the skeleton sequences and recombines the elements to generate a new skeleton sequence, which can be viewed at any desired view-angle.
We implement the action transfer based on Transmomo \cite{yang2020transmomo} without using any paired data for supervision. The transfer network is trained in an unsupervised manner by exploiting invariance properties of three orthogonal factors of variation including motion, structure, and view-angle. The motion is invariant despite structural and view-angle perturbations, the structure is consistent through time and invariant despite view-angle perturbations, and the view-angle is consistent through time and invariant despite structural perturbations.

\begin{figure*}
\begin{center}
\subfigure[Trajectories before motion transfer from student 1.]{
\begin{minipage}[t]{0.4\linewidth}
\centering
\includegraphics[height=4.3cm]{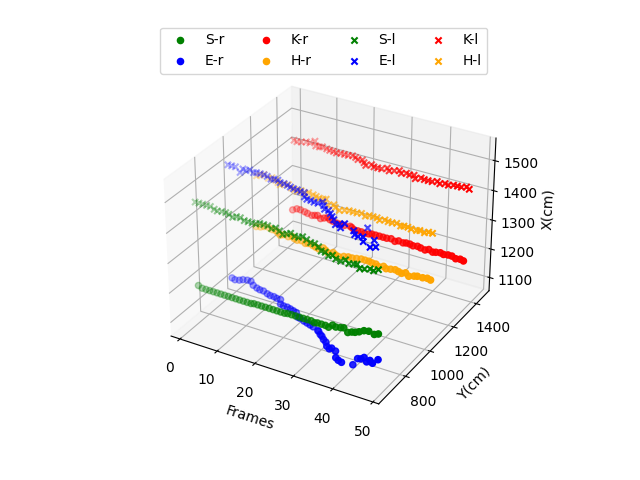}
\end{minipage}
}
\subfigure[Trajectories after motion transfer from student 1.]{
\begin{minipage}[t]{0.4\linewidth}
\centering
\includegraphics[height=4.3cm]{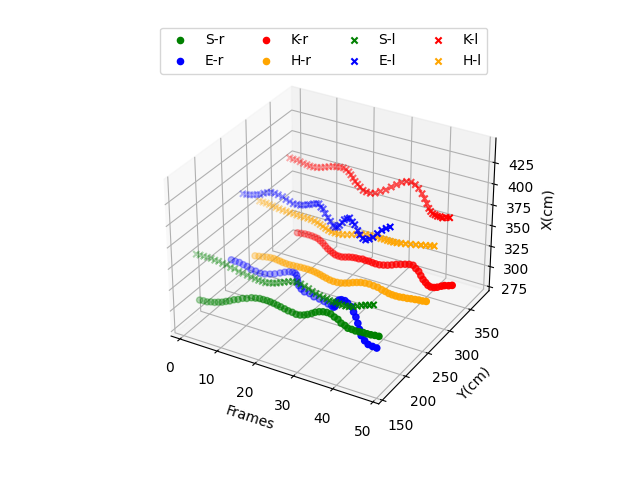}
\end{minipage}
}
\subfigure[Trajectories before motion transfer from student 2.]{
\begin{minipage}[t]{0.4\linewidth}
\centering
\includegraphics[height=4.3cm]{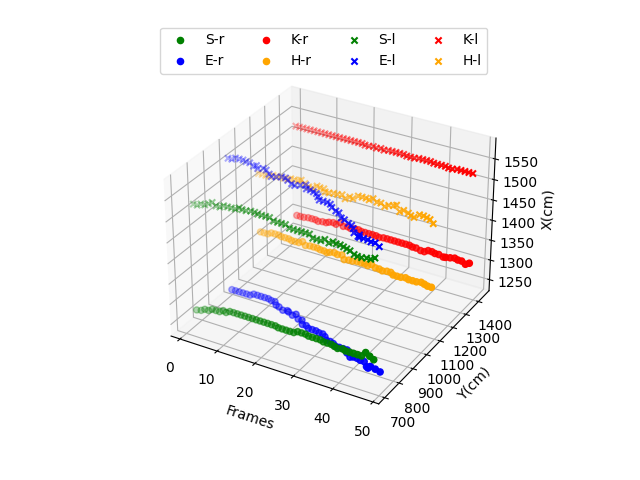}
\end{minipage}
}
\subfigure[Trajectories after motion transfer from student 2.]{
\begin{minipage}[t]{0.4\linewidth}
\centering
\includegraphics[height=4.3cm]{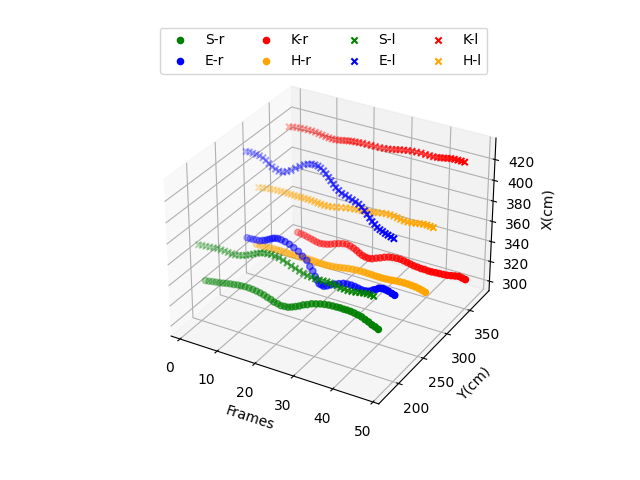}
\end{minipage}
}
\end{center}
  \caption{The changes of the key joints' trajectories before and after motion transfer.}
\label{fig:travis}
\end{figure*}

\begin{figure*}
\begin{center}
\subfigure[Angles before motion transfer from student 1.]{
\begin{minipage}[t]{0.4\linewidth}
\centering
\includegraphics[height=4.3cm]{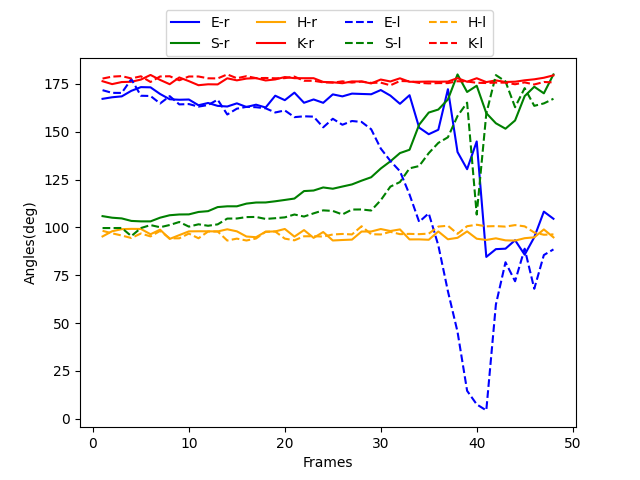}
\end{minipage}
}
\subfigure[Angles after motion transfer from student 1.]{
\begin{minipage}[t]{0.4\linewidth}
\centering
\includegraphics[height=4.3cm]{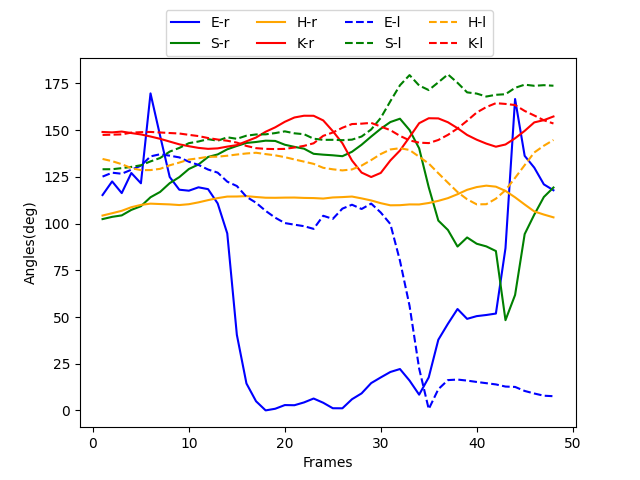}
\end{minipage}
}
\subfigure[Angles before motion transfer from student 2.]{
\begin{minipage}[t]{0.4\linewidth}
\centering
\includegraphics[height=4.3cm]{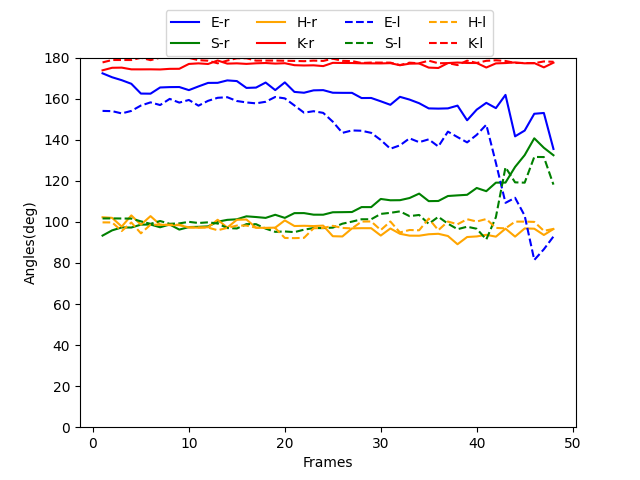}
\end{minipage}
}
\subfigure[Angles after motion transfer from student 2.]{
\begin{minipage}[t]{0.4\linewidth}
\centering
\includegraphics[height=4.3cm]{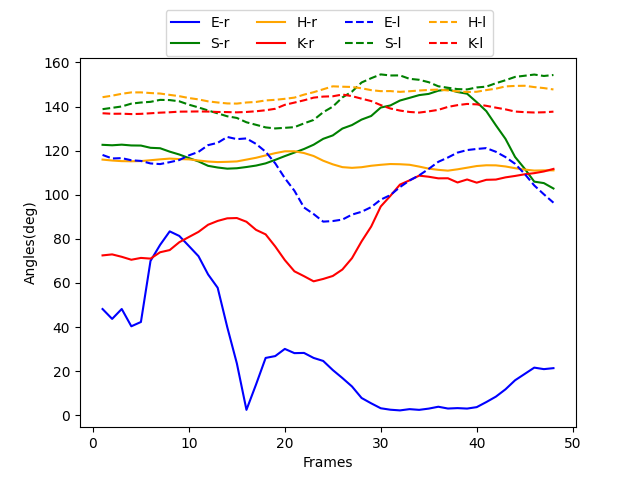}
\end{minipage}
}
\end{center}
  \caption{The changes of the key joints' angles before and after motion transfer.}
\label{fig:anglevis}
\end{figure*}
For an input sequence $\mathbf{s}^\mathrm{T}\in \mathbb{R}^{\mathrm{T}\times2\mathrm{N}}$ where $\mathrm{T}$ is the length of the skeleton sequence and $\mathrm{N}$ is the number of body joints.
The motion encoder uses several layers of one dimensional temporal convolution to extract the motion information. The structure encoder has a similar network structure with the difference that the final structure code is obtained after a temporal max pooling. The view code is obtained the same way we obtained the structure code.
The decoder takes the motion, body and view codes as input and reconstructs a 3D joint sequence $\mathbf{S}^\mathrm{T}\in \mathbb{R}^{\mathrm{T}\times3\mathrm{N}}$ through convolution layers, in symmetry with the encoders.
The total loss function is derived based on invariance and weighted by the following loss terms:

\begin{equation}\label{}
\begin{split}
\mathbf{L}_{total}=\lambda_{rec} \mathbf{L}_{rec}+\lambda_{crs} \mathbf{L}_{crs}+\lambda_{trip} \mathbf{L}_{trip} \\
+\lambda_{inv} \mathbf{L}_{inv}+\lambda_{adv} \mathbf{L}_{adv},
\end{split}
\end{equation}
where $\mathbf{L}_{rec}$ is the reconstruction loss to minimize the difference between real data and 3D reconstructions projected back to 2D, $\mathbf{L}_{crs}$ is the cross reconstruction loss for two sequences, $\mathbf{L}_{trip}$ is the triplet loss to map views, $\mathbf{L}_{inv}$ is the structural invariance loss to ensure that the view code is invariant to structural change estimations from the same sequence to a small neighborhood while alienating estimations from rotated sequences, and $\mathbf{L}_{adv}$ is used to measure the domain discrepancy between the projected 2D sequences and real 2D sequences. 

We evaluate the action quality by comparing the trajectory and angle changes of the key joints, such as \emph{shoulder}, \emph{elbow}, \emph{hip} and \emph{knee}. We select the first 15 key points defined in Body-25 for motion transfer. For the $j$-th joint in the skeleton model, the angle sequence ($\theta_j^\mathrm{T}$) for an action sample is calculated as follows:
\begin{equation}\label{}
\theta_j^\mathrm{T}=\arccos(\frac{(\mathbf{S}_{j-1} -\mathbf{S}_{j})\cdot (\mathbf{S}_{j+1} -\mathbf{S}_{j})}{\left | \mathbf{S}_{j-1} -\mathbf{S}_{j}\left |  \right | \mathbf{S}_{j+1} -\mathbf{S}_{j} \right | } ),j\in(0,15)
\end{equation}

\begin{figure}
\begin{center}
 \includegraphics[height=4.2cm]{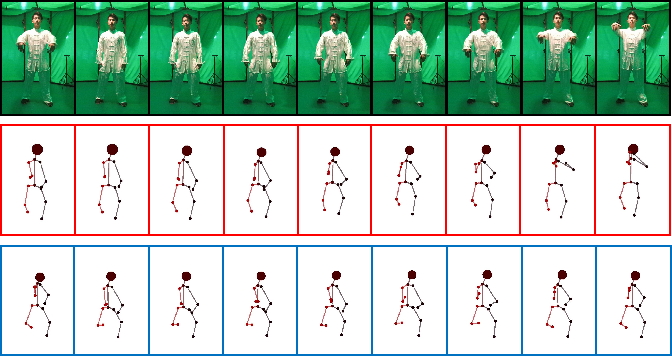}
\end{center}
  \caption{Examples of the re-targeted key-frame skeletons (top: the coach; middle: student 1; bottom: student 2).}
\label{fig:ret}
\end{figure}

In our experiments, we take the RGB image (1920$\times$1080p) sequence of the coach collected by Kinect camera as the target, and two skeleton sequences of the students captured by our multi-camera system as input data.
Figure \ref{fig:travis} and Figure \ref{fig:anglevis} show the changes of the trajectories and angels of the \emph{shoulder} (S-l and S-r), \emph{elbow} (E-l and E-r), \emph{hip} (H-l and H-r) and \emph{knee} (K-l and K-r) joints respectively. The length of the action sample is 48 frames. By analyzing the changes of trajectories and angles after motion transfer, we can find the major difference joints and emergence moments. In the case of these two students, the differences of them are mainly concentrated in the \emph{shoulder} and \emph{knee} joints during the movements located at the end of the action.
Figure \ref{fig:ret} shows the visual analysis results of re-targeted skeletons from two students compared with the coach. It can be seen that the TaiChi motion of student1 is more standard. According to the scores of professional coach, student 1 (100) also scored higher than student 2 (86).

\section{Conclusions}

In this paper, we build a multi-camera system and realize TaiChi performance capture and analysis with multi-view geometry and artificial intelligence technology.
Traditional visual method and deep-learning base NeRFs are used in combination to achieve sparse 3D skeleton and dense 3D surface reconstruction. we collect and organize the TaiChi data with multi RGB cameras, and process them to experimental analysis. To realize TaiChi performance analysis for different groups, skeleton sequences are normalized modeled with motion transfer and then further assessed with the changes of the joints' trajectories and angles. We also carry out evaluation experiments and the experimental results have shown the efficiency of our system.
In the future, we will research more accurate and robust analysis methods with data from less camera views.

\section*{Acknowledgments}
The work is assisted by Haiqing Hu, Jinyang Li, Xinyu Wang and Tianhan Zhang from Beijing Sports University, who help us collect and process the data.
This work is partially supported by the National Key R$\&$D Program of China (No. 2022YFC3600300, No. 2022YFC3600305), and the Fundamental Research Funds for Central Universities No.2022QN018.

\bibliographystyle{named}
\bibliography{ijcai23}

\end{document}